\documentclass[10pt,twocolumn,letterpaper]{article}

 \usepackage[pagenumbers]{cvpr} 

\usepackage{mathtools}
\usepackage[dvipsnames]{xcolor}
\usepackage{pifont}
\newcommand{\cmark}{{\color{Green}\ding{51}}}%
\newcommand{\xmark}{{\color{Red}\ding{55}}}%
\usepackage{multirow}

\usepackage[dvipsnames]{xcolor}

\definecolor{cvprblue}{rgb}{0.21,0.49,0.74}
\usepackage[pagebackref,breaklinks,colorlinks,citecolor=cvprblue]{hyperref}

\def\paperID{64} 
\def\confName{3DV\xspace}
\def\confYear{2025\xspace}

\title{Alignment Scores: Robust Metrics for \\Multiview Pose Accuracy Evaluation
}

\author{Seong Hun Lee\thanks{This work was supported by the Spanish
Government (projects PID2021-127685NB-I00 and TED2021-131150BI00) and the Arag\'on Government
(project DGA T45\_23R).}
\hspace{4em}  Javier Civera\\
I3A, University of Zaragoza\\
Mar\'{i}a de Luna 1, 50018 Zaragoza, Spain\\
{\tt\small seonghunlee@unizar.es \hspace{1em} jcivera@unizar.es}
}

\begin{document}
\maketitle

\begin{abstract}
We propose three novel metrics for evaluating the accuracy of a set of estimated camera poses given the ground truth:
Translation Alignment Score (TAS), Rotation Alignment Score (RAS), and Pose Alignment Score (PAS).
The TAS evaluates the translation accuracy independently of the rotations, and the RAS evaluates the rotation accuracy independently of the translations.
The PAS is the average of the two scores, evaluating the combined accuracy of both translations and rotations.
The TAS is computed in four steps: 
(1) Find the upper quartile of the closest-pair-distances, $d$.
(2) Align the estimated trajectory to the ground truth using a robust registration method.
(3) Collect all distance errors and obtain the cumulative frequencies for multiple thresholds ranging from $0.01d$ to $d$ with a resolution $0.01d$.
(4) Add up these cumulative frequencies and normalize them such that the theoretical maximum is 1.
The TAS has practical advantages over the existing metrics in that (1) it is robust to outliers and collinear motion, and (2) there is no need to adjust parameters on different datasets.
The RAS is computed in a similar manner to the TAS and is also shown to be more robust against outliers than the existing rotation metrics.
We verify our claims through extensive simulations and provide in-depth discussion of the strengths and weaknesses of the proposed metrics.
\end{abstract}

\begin{table*}[t]
\small
\begin{center}
\begin{tabular}{ccccc}
\hline
 & Metric scale  & Robust to & Robust to & Insensitive to the \\
 & unnecessary? & outliers? & Collinearity  & trajectory length? \\
\hline
\multicolumn{1}{c}{Absolute trajectory error (ATE) \cite{sturm_2012_benchmark}} & \xmark & \xmark  & \cmark &\cmark\\
\hline
\multicolumn{1}{c}{Mean average accuracy (mAA) \cite{yi_2018_cvpr, jin_2021_ijcv}} & \cmark & \cmark  & \xmark &\xmark \\
\hline
\multicolumn{1}{c}{Discernible trajectory \& rotation error (DTE \& DRE) \cite{dte} } & \cmark  & \cmark & \cmark & \xmark \\
\hline
\multicolumn{1}{c}{Translation, Rotation \& Pose Alignment Score (TAS, RAS \& PAS)} & \cmark  &\cmark& \cmark & \cmark\\
\hline
\end{tabular}
\end{center}
\caption{
Qualitative comparison of various metrics for camera trajectory evaluation given the ground truth.
To the best of our knowledge, our metrics are the only ones that tick all the boxes.
}
\label{tab:method_comparison}
\end{table*}

\section{Introduction}
\label{sec:intro}
Estimating a set of unknown camera poses from images (and optionally other sensors) is a fundamental problem in 3D vision and robotics.
This problem plays a crucial role in many research domains, including 
odometry \cite{dso, leutenegger_2015_keyframe, loam}, simultaneous localization and mapping (SLAM) \cite{lcsd, ORBSLAM3_TRO, droid}, visual localization \cite{sattler_2018_benchmarking, lynen_2020_ijrr, toft_2022_long} and structure-from-motion (SfM)  \cite{agarwal_2009_iccv, sfm_revisited, moulon_2013_iccv}.
All of these topics are highly relevant to applications such as photogrammetry, autonomous robot navigation, AR/VR and 3D rendering.

A fair and reliable evaluation of the multiview pose estimation results is essential, as it allows for the characterization of the absolute and relative performance of both existing and newly proposed methods. 
This evaluation includes several key aspects, such as the availability of representative and well-curated datasets (\textit{e.g.}, see \cite{kitti,wilson20141dsfm,burri2016euroc,knapitsch2017tanks,schubert2018tum}) and the choice of appropriate evaluation metrics. 
Both of these aspects are equally important, but the latter is frequently overlooked.

When the ground-truth pose data is available\footnote{When this ground-truth data is not available, a commonly used metric is the reprojection error \cite{hartley_book, sfm_revisited}.
In this work, we focus on the case where the ground-truth poses are available for the evaluation.
Therefore, we omit the discussion of the reprojection error.}, one can evaluate the accuracy of the results by aligning the estimated poses to the ground truth and looking at the difference between them.
At first glance, coming up with an appropriate performance metric may seem straightforward, but it turns out to be a complex problem once we consider the following desiderata:

\begin{enumerate}
    \item It should not require the ground-truth position data to be in metric units.
    This is because the ground truth may be either generated synthetically (for simulations) or obtained using an existing SfM system (\textit{e.g.}, 1DSfM \cite{wilson20141dsfm} and MVS \cite{zhou_2018_deeptam}).
    \item It should be robust to outliers.
    To be more specific, in the presence of multiple outliers, the metric should respond discernibly to the varying amount of outliers and inlier estimation error (\textit{i.e.}, noise).
    \item It should not be affected by collinear camera motions, which is common when the dataset is collected on a wheeled vehicle (\textit{e.g.}, KITTI dataset \cite{kitti}).
    \item It should not be affected by the trajectory length.
    In other words, the metric should neither reward nor penalize a method based solely on the number of cameras in the ground-truth data.
\end{enumerate}
In practice, these criteria should be taken into consideration for the evaluation metric to be robust and versatile across different types of datasets and estimation methods.
To our knowledge, however, none of the existing metrics in the literature meets all of the these criteria (see Tab. \ref{tab:method_comparison}).

In this work, we propose novel metrics that meet the aforementioned criteria.
Our contributions are as follows:

\begin{itemize}
    \item We propose Translation Alignment Score (TAS) that evaluates the camera translation accuracy independently of the rotations.
    We use the ground-truth distances between the cameras to obtain the error threshold, which we then use to evaluate the closeness between the ground-truth and estimated camera positions after robust alignment.
    Specifically, we evaluate the sum of the cumulative frequencies for multiple error thresholds, as in \cite{yi_2018_cvpr, jin_2021_ijcv}.
    We show that this metric is robust to outliers, collinear motion, and the change in the trajectory length.
    \item We propose Rotation Alignment Score (RAS) that evaluates the camera rotation accuracy independently of the translations.
    Similar to the TAS, the RAS is also computed using the cumulative frequencies for multiple error thresholds and is robust to a large number of outliers.
    \item We propose Pose Alignment Score (PAS) as the average of the TAS and RAS. 
    This metric can be used as a single metric evaluating the combined accuracy of translations and rotations.
\end{itemize}

We perform an extensive evaluation of the proposed metrics in comparison with the existing ones. 
Our evaluation reveals previously unknown limitations of the existing metrics, as well as the practical advantages of ours.
    
The paper is organized as follows:
In Section \ref{sec:related}, related work is discussed.
We describe our metrics, TAS, RAS and PAS, in Section \ref{sec:tas}, \ref{sec:ras} and \ref{sec:pas} respectively.
In Section \ref{sec:evaluation}, we present the evaluation results.
We summarize the findings in Section \ref{sec:summary} and provide additional discussions in Section \ref{sec:discussion}.
Finally, conclusions are presented in Section \ref{sec:conclusion}.

Our code is publicly available at \url{https://github.com/sunghoon031/AlignmentScores}.

\section{Related Work}
\label{sec:related}
In the early work by Sturm et al. \cite{sturm_2012_benchmark}, the absolute trajectory error (ATE) and the relative pose error (RPE) are compared and discussed in the context of visual odometry/SLAM evaluation.
Unlike the RPE that is applicable only to odometry methods (\textit{e.g.}, \cite{kummerle_2009_auro, burgard_2009_iros, geiger_2012_cvpr}), the ATE can also be used to evaluate the reconstruction from a set of unordered images.
Zhang and Scaramuzza \cite{zhang_2018_iros} provide further analysis of these two error metrics for visual(-inertial) pipelines.

Lee and Civera \cite{dte} show that one of the main problems of the ATE is that it is highly sensitive to outliers.
A single extreme outlier can easily render the ATE useless, as it becomes incapable of discerning the varying inlier trajectory accuracy or the number of outliers.
They propose alternative error metrics, called the discernible trajectory error (DTE) and discernible rotation error (DRE).
The key idea behind these metrics is to align the estimated trajectory to the ground truth using robust $L_1$ optimization methods.
They also winsorize and normalize the errors to limit the influence of unbounded outliers. 
This step, however, involves (i) a threshold which depends on the dimension of the ground-truth trajectory, and (ii) a manually set parameter which may need to be tuned per dataset.

Yi et al. \cite{yi_2018_cvpr} propose another metric that is also robust to outliers.
This metric is originally called mean average precision (mAP), but later renamed as mean average accuracy (mAA) by Jin et al. \cite{jin_2021_ijcv} for correct terminology.
Unlike the ATE and DTE, the mAA is a metric based solely on angular errors.
To compute this metric, they first collect the angular differences between the estimated and ground-truth translation and rotation vectors between every possible pair of cameras.
Then, the mAA is obtained as the area under the normalized cumulative histogram of these angular differences.
For reference, the maximum threshold for this histogram is set to $10^\circ$ in \cite{jin_2021_ijcv}.
One of the limitations of this metric is that it is prone to degeneracy when the camera motion is collinear.

In this work, we draw inspiration from the DTE/DRE \cite{dte} and the mAA \cite{yi_2018_cvpr, jin_2021_ijcv} and do the following:

\begin{enumerate}
    \item Similar to the DTE/DRE, we robustly align the camera trajectories/rotations in order to compute the difference between the estimation and ground truth. 
    In contrast to the DTE/DRE, however, we completely decouple the translations and the rotations.
    Also, our method does not involve a threshold that depends on the length of the ground-truth trajectory.
    \item Similar to the mAA, we use the cumulative frequency histogram of the errors.
    In contrast to the mAA, however, we do not use the angular errors between every possible camera pair.
    Instead, we use distance errors between the estimated and ground-truth camera trajectories to evaluate the translation accuracy.
    Furthermore, rather than combining the translation and rotation errors into one cumulative histogram, we use two separate histograms, one for translations and the other for rotations.
    This allows us to evaluate the accuracy of the translations and rotations separately (if we want).
\end{enumerate}

\section{Translation Alignment Score (TAS)}
\label{sec:tas}
\begin{figure}[t]
 \centering
 \includegraphics[width=0.47\textwidth]{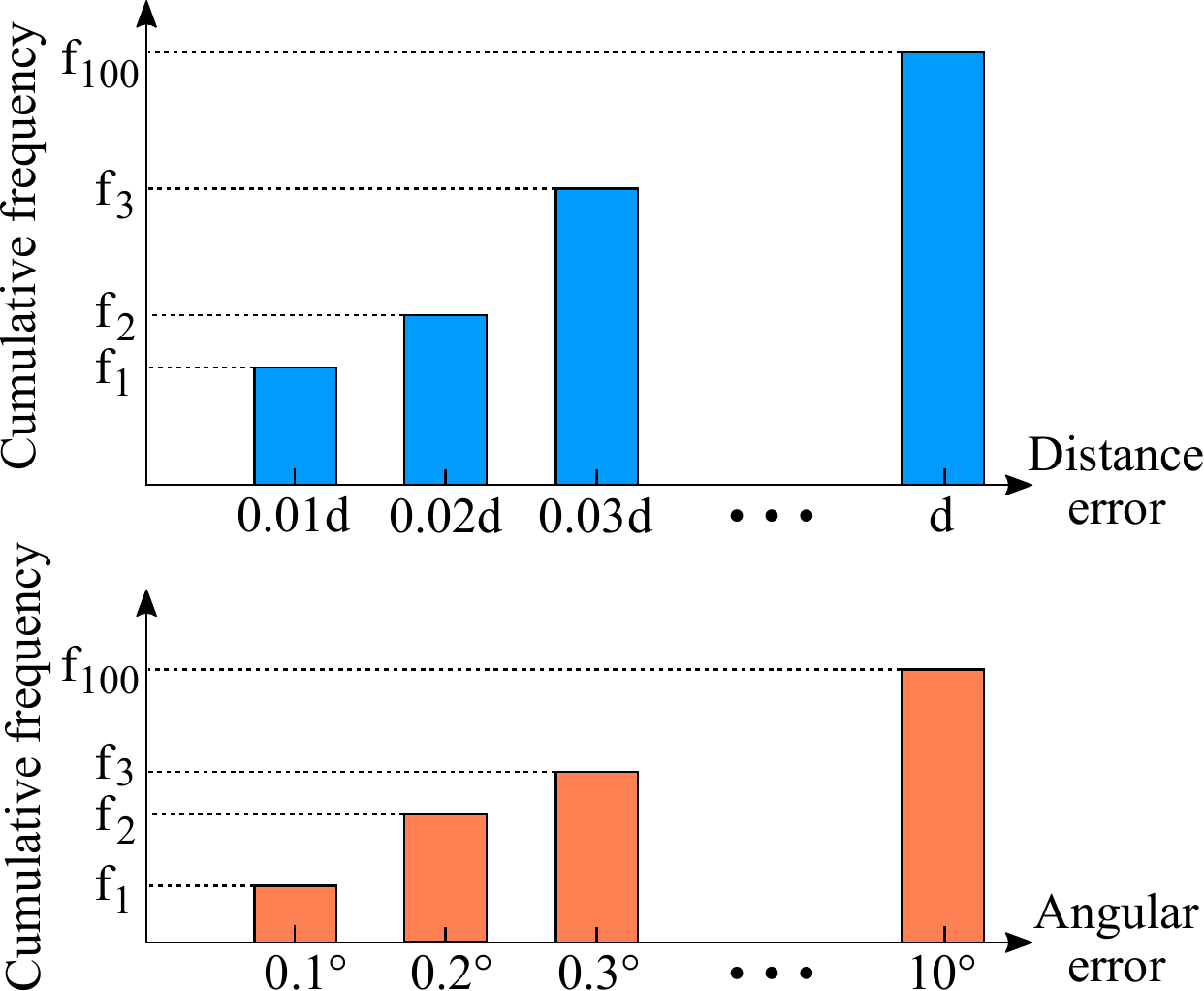}
\caption{\textbf{[Top]} Cumulative frequency histogram of the distance errors between the aligned camera positions and the ground truth.
\textbf{[Bottom]} Cumulative frequency histogram of the angular errors between the aligned camera rotations and the ground truth.
Note that the cumulative frequencies in these two histograms are not necessarily the same.
}
\label{fig:histogram}
\end{figure}

The TAS is computed as follows:

\begin{enumerate}
    \item For each of the $n$ cameras, find the distance to the nearest camera.
    Let $d$ be the upper quartile of these $n$ closest-pair-distances, \textit{i.e.,}
    \begin{equation}
    \label{eq:d}
        d = \underset{i\in\mathcal{C}}{\mathrm{Q3}} \left(\min_{j \in\mathcal{C}, j\neq i} \lVert\mathbf{c}_i-\mathbf{c}_j\rVert\right),
    \end{equation}
    where $\mathcal{C}=\{1, 2, \cdots, n\}$ and $\mathbf{c}_i$ represents the $i$-th ground-truth camera position.
    \item Align the estimated camera positions to the ground truth using a robust point cloud registration method. 
    The TAS is agnostic about the registration method as long as it is sufficiently robust to outliers.
    In this work, we use PCR-99 \cite{pcr99}, modified such that the registration can be done without requiring the prior knowledge of the inlier threshold. 
    Specifically, we modify the original unknown-scale version of PCR-99 as follows: 
    \begin{enumerate}
        \item Instead of the sample ordering (Section IV.B in \cite{pcr99}), do random sampling.
        \item Collect the first 1000 hypotheses that pass the prescreening test (as proposed in \cite{pcr99}).
        \item Among these 1000 hypotheses, choose the one that produces the smallest cost. 
        The cost is defined as the $m$-th smallest distance error between the aligned camera positions and the ground truth, where 
        \begin{equation}
            m = \max\left(4, \mathrm{round}(n/10)\right).
        \end{equation}
    \end{enumerate}

    \item Collect all distance errors between the aligned and the ground-truth trajectories.
    Obtain the cumulative frequencies for multiple error thresholds.
    We set the thresholds to $\{0.01d, 0.02d, \cdots, d\}$ (see the top plot of Fig. \ref{fig:histogram}).

    \item Finally, compute the TAS by adding up these cumulative frequencies and normalizing it such that the theoretical maximum is 1, \textit{i.e.,}
    \begin{equation}
        \mathrm{TAS} = \frac{1}{100n}\left(\sum_{k=1}^{100} f_k\right),
    \end{equation}
    where $f_k$ corresponds to the $k$-th cumulative frequency, as shown in the top plot of Fig. \ref{fig:histogram}.
\end{enumerate}

\noindent\textbf{Novel contributions:}
First of all, the key idea of our method is to set the error thresholds based on the upper quartile of the closest-pair-distances, $d$ in \eqref{eq:d}.
For example, consider a dataset that has 10 or more cameras.
Suppose more cameras are added in this dataset (\textit{e.g.,} as the coverage of the scene grows wider).
In this case, the $d$ value is unlikely to change significantly, unless the overall spacing between the cameras increases drastically.
This makes the TAS fairly insensitive to the change in the trajectory length.
In contrast, the DTE \cite{dte} normalizes the errors using the median absolute deviations (MAD) of the ground-truth camera positions, which makes it dependent on the trajectory length.
This difference will be demonstrated in Section \ref{subsec:result_size}.
Second, since we consider the distance errors, rather than angular errors (as in the mAA \cite{yi_2018_cvpr, jin_2021_ijcv}), our metric can handle collinear motion.
This will be shown in Section \ref{subsec:result_collinear}.
Finally, the TAS can be obtained even when the ground-truth rotations are unavailable, which is not the case for the mAA \cite{yi_2018_cvpr, jin_2021_ijcv} and DTE \cite{dte}.

\section{Rotation Alignment Score (RAS)}
\label{sec:ras}
The RAS is computed using a similar method to the TAS:

\begin{enumerate}
    \item Align the estimated camera rotations to the ground truth using a robust single rotation averaging method.
    The RAS is agnostic about the averaging method as long as it is sufficiently robust to outliers.
    In this work, we use \cite{sra2}.
    \item Collect all angular errors between the aligned and the ground-truth rotations. 
    Obtain the cumulative frequencies for multiple error thresholds.
    We set the thresholds to $\{0.1^\circ, 0.2^\circ, \cdots, 10^\circ$\} (see the bottom plot of Fig. \ref{fig:histogram}).
    \item Compute the RAS by adding up these cumulative frequencies and normalizing it such that the theoretical maximum is 1, \textit{i.e.,}
    \begin{equation}
        \mathrm{RAS} = \frac{1}{100n}\left(\sum_{k=1}^{100} f_k\right),
    \end{equation}
    where $f_k$ corresponds to the $k$-th cumulative frequency, as shown in the bottom plot of Fig. \ref{fig:histogram}.
\end{enumerate}

\noindent Note that the RAS can be computed without the knowledge of the ground-truth translations.
This means that it can be used for rotation-only estimations, such as multiple rotation averaging \cite{chatterjee_2018_tpami, lee_2022_cvpr} and rotation-only bundle adjustment \cite{lee_2021_cvpr}.

\section{Pose Alignment Score (PAS)}
\label{sec:pas}
While the TAS and RAS can be used together to represent the translation and rotation accuracy, it may also be desirable to have a single metric that represents the six degrees-of-freedom pose accuracy as a whole.
To this end, we propose another metric called Pose Alignment Score (PAS), which is simply the average of the TAS and RAS:
\begin{equation}
\label{eq:pas}
    \mathrm{PAS}=\frac{\mathrm{TAS}+\mathrm{RAS}}{2}.
\end{equation}
As both TAS and RAS are bounded between 0 and 1, PAS also takes a value between 0 and 1.
As will be shown in Section \ref{sec:result_pas}, the PAS is a robust metric that can discern the varying noise level in both translations and rotations, and it can do so more consistently than the mAA \cite{yi_2018_cvpr, jin_2021_ijcv}.

\section{Evaluation}
\label{sec:evaluation}

\begin{figure}[t]
 \centering
 \includegraphics[width=0.47\textwidth]{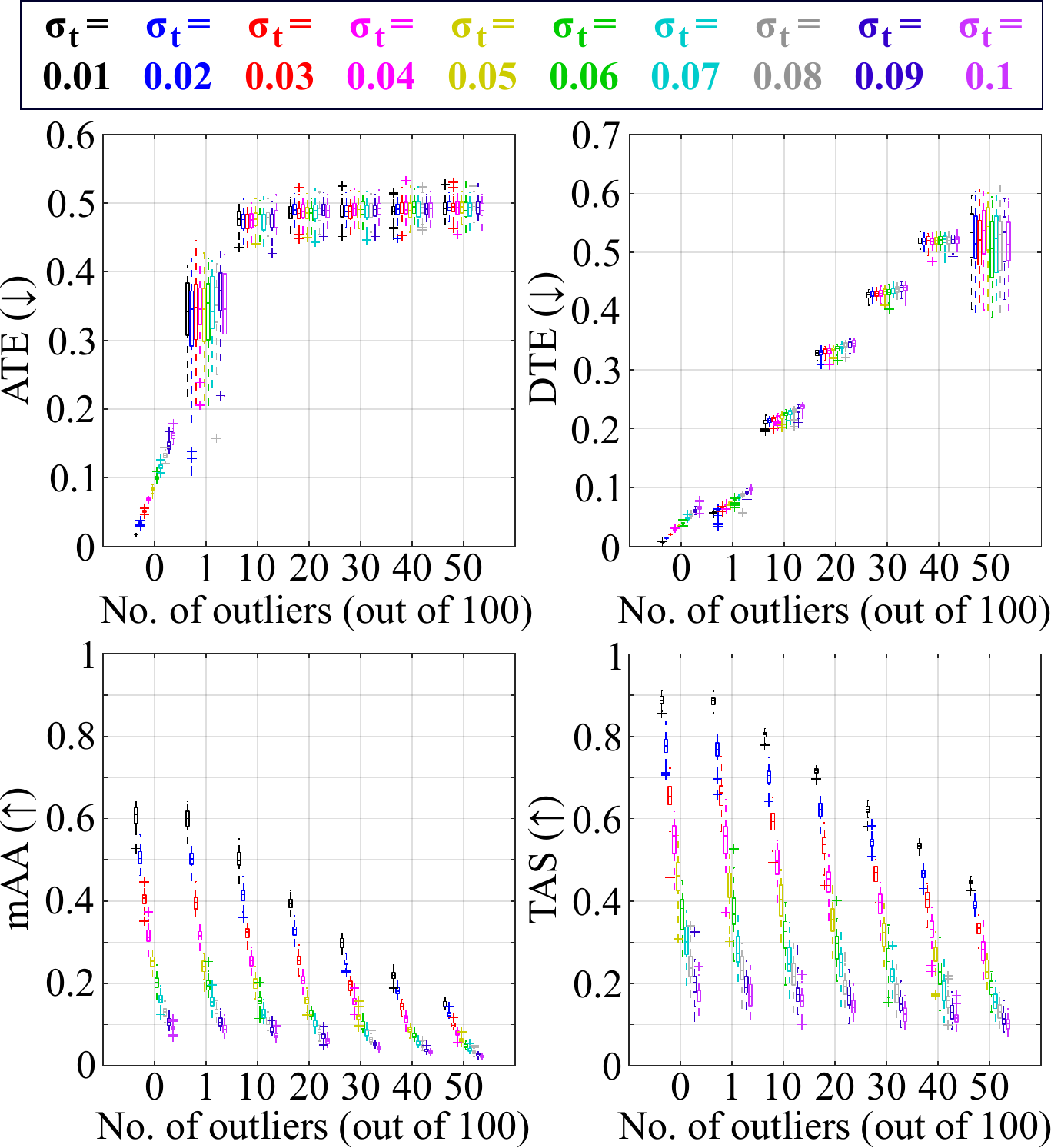}
\caption{\textbf{[Random translations]} Comparison of the four translation metrics under different noise levels and numbers of outliers.
The ATE is the most sensitive to outliers: 
When we add a single outlier in the estimation, it immediately loses its power to discern the varying noise level.
The DTE is more robust than the ATE, but the mAA and TAS have stronger discerning power across a wider range of outliers.
Comparing the mAA and TAS in terms of the sensitivity to the noise level, the latter is shown to have more consistent sensitivity across the range of outliers.
For instance, the range of the mAA at 50 outliers is 74\% smaller than that at zero outliers, while it is 51\% smaller for the TAS.
}
\label{fig:results_outliers}
\end{figure}

\begin{figure}[t]
 \centering
 \includegraphics[width=0.47\textwidth]{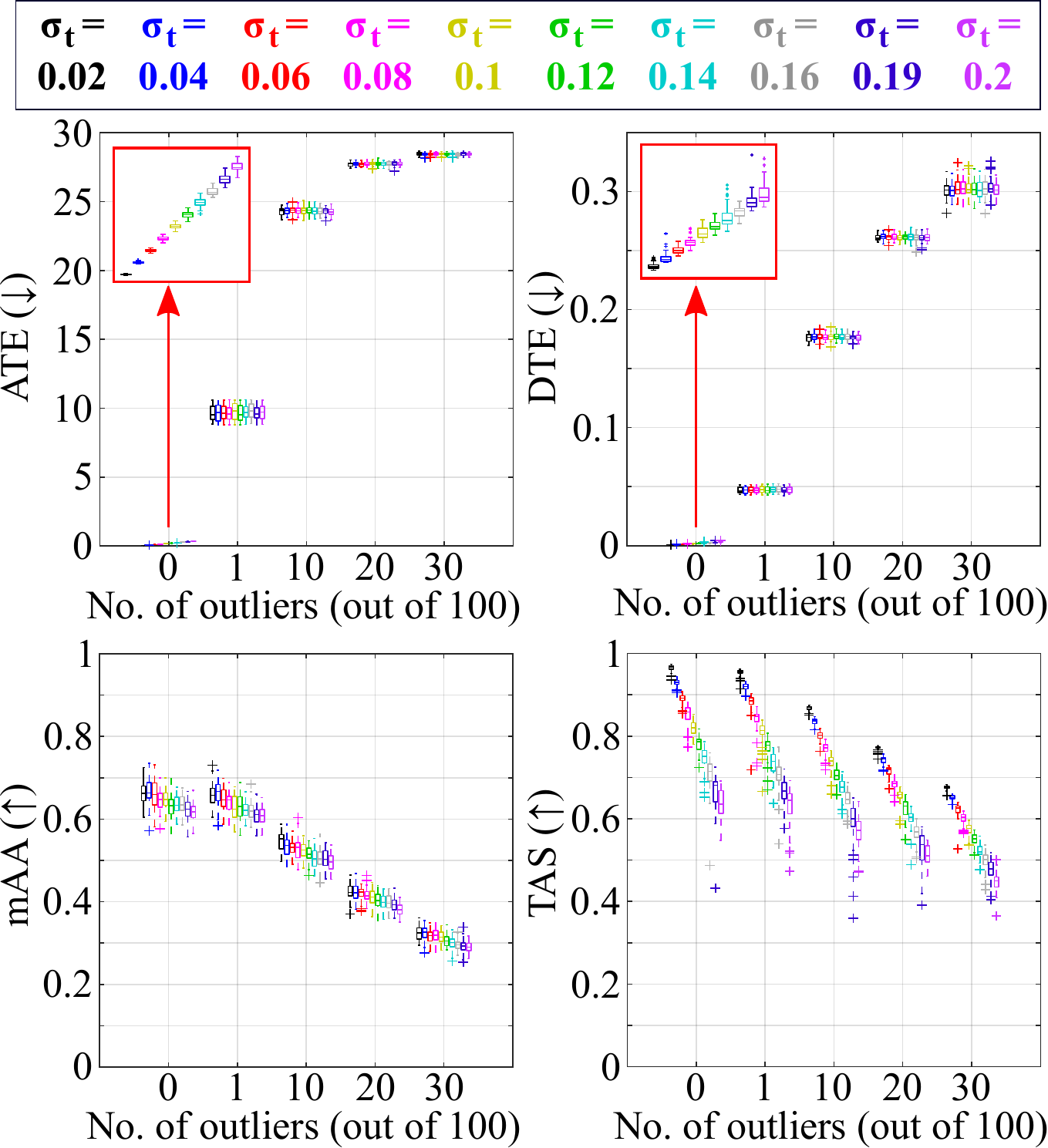}
\caption{\textbf{[Collinear translations]} Comparison of the four translation metrics under different noise levels and numbers of outliers.
The ATE and DTE can handle collinear motion only in the absence of outliers: 
When we add a single outlier in the estimation, both metrics lose their power to discern the varying noise level.
The mAA has weak discerning power, with or without outliers. 
The TAS is the only metric that maintains strong discerning power as the amount of noise and outliers varies.}
\label{fig:results_collinear}
\end{figure}

To evaluate the proposed metrics, we use carefully designed simulations instead of running an existing estimation method on real-world datasets.
This allows us to conduct a controlled study of various factors such as the estimation noise, collinearity, and the number of views and outliers, which is otherwise impossible.
Since we want to test the metrics to their limits, while being agnostic about the pose estimation method, we choose to perform all our evaluations using meticulously simulated ground-truth and estimation data only.\footnote{This choice is made despite the fact that our simulation may not be the most realistic depiction of the commonly encountered scenarios in certain applications such as large-scale odometry (which is prone to drift) and kinematically constrained motion estimation.
We argue that this trade-off is (i) inevitable for the purpose of this work, and (ii) justifiable, since the evaluation is done stochastically in both generic and specific scenarios that are not entirely unrealistic in the real world.
The study of different types of noise and outlier distributions is beyond the scope of this work.}

\subsection{Evaluation of TAS}
\label{sec:result_tas}
In this section, we compare the TAS with three other metrics for camera trajectory evaluation:

\begin{enumerate}
    \item Absolute trajectory error (ATE) \cite{sturm_2012_benchmark},
    \item Discernible trajectory error (DTE)
    \cite{dte},
    \item Mean average accuracy (mAA) \cite{yi_2018_cvpr, jin_2021_ijcv}.
\end{enumerate}

\subsubsection{Robustness to outliers}
\label{subsec:result_outlier}

To evaluate the robustness to outliers, we run Monte Carlo simulations, set up as follows:
We generate 100 ground-truth cameras with random rotations and positions inside a unit cube centered at the origin.
Then, we obtain the estimated trajectory by perturbing the camera positions with Gaussian noise $\mathcal{N}(0, \sigma_t^2)$, $\sigma_t\in\{0.01, 0.02, \cdots, 0.1\}$, and the rotations with $\sigma_r=3^\circ$.
Some of the cameras are turned into outliers with uniformly distributed random rotations and positions inside a $10^3$ cube centered at the origin. 
We repeat this procedure to produce 50 independent trajectories for each different setting of the noise level $\sigma_t$ and the number of outliers.
The results and their interpretation are shown in Figure \ref{fig:results_outliers} and its caption, respectively.

\begin{figure}[t]
 \centering
 \includegraphics[width=0.47\textwidth]{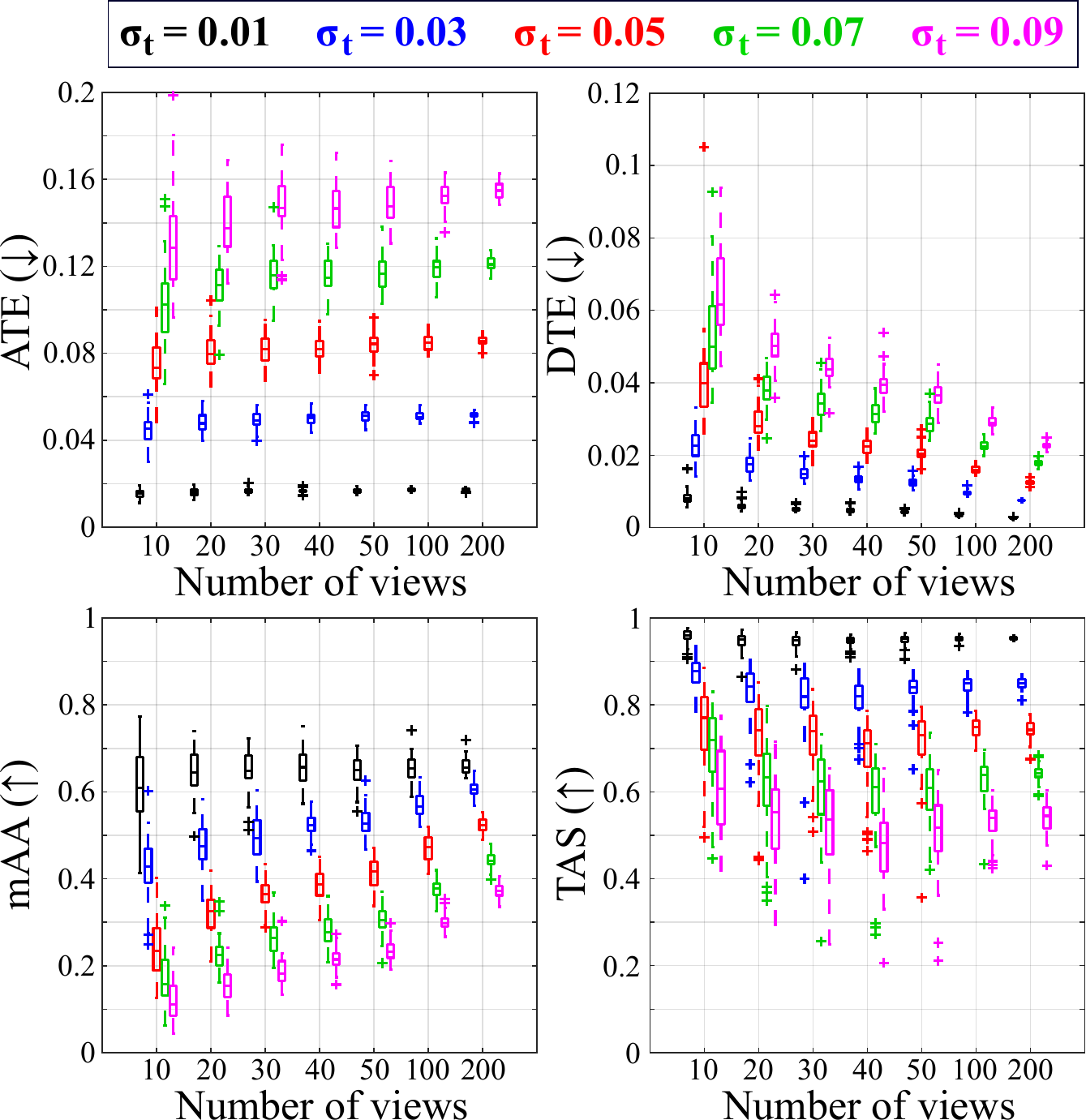}
\caption{\textbf{[Random translations]} Comparison of the four translation metrics under different noise levels and trajectory lengths.
The DTE and mAA tend to favor larger datasets when the noise level is the same.
The ATE and TAS are much less sensitive to the varying trajectory length than the other two, especially at moderate noise levels.}
\label{fig:results_volume}
\end{figure}

\begin{figure*}[t]
 \centering
 \includegraphics[width=\textwidth]{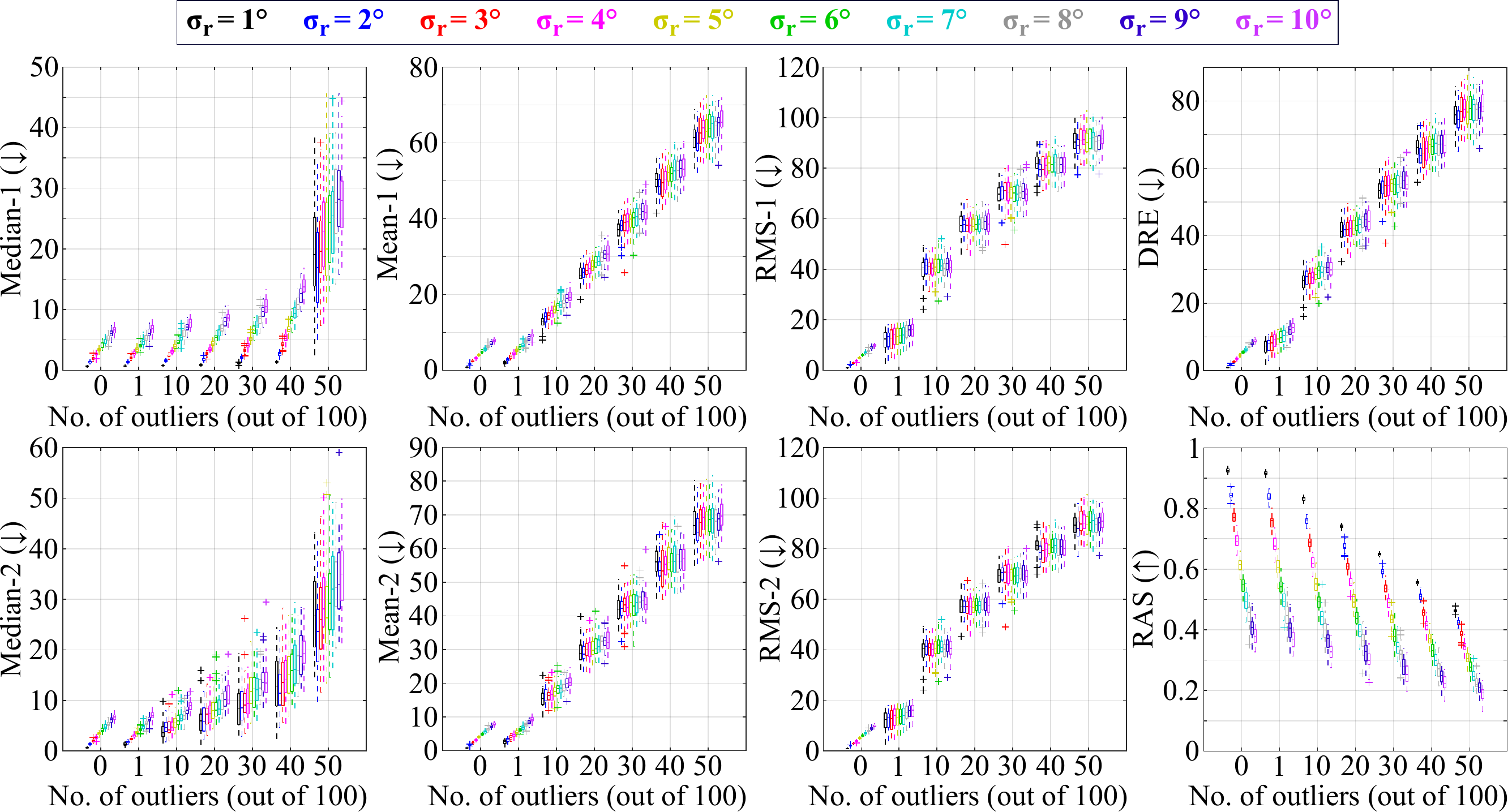}
\caption{\textbf{[Rotations]} Comparison of the eight rotation metrics under different noise levels and numbers of outliers.
The RAS has the strongest discerning power when the amount of noise and outliers varies.
In other words, it can discern different noise levels when the number of outliers is fixed, and vice versa.}
\label{fig:results_rotation}
\end{figure*}

\begin{figure*}[t]
 \centering
 \includegraphics[width=\textwidth]{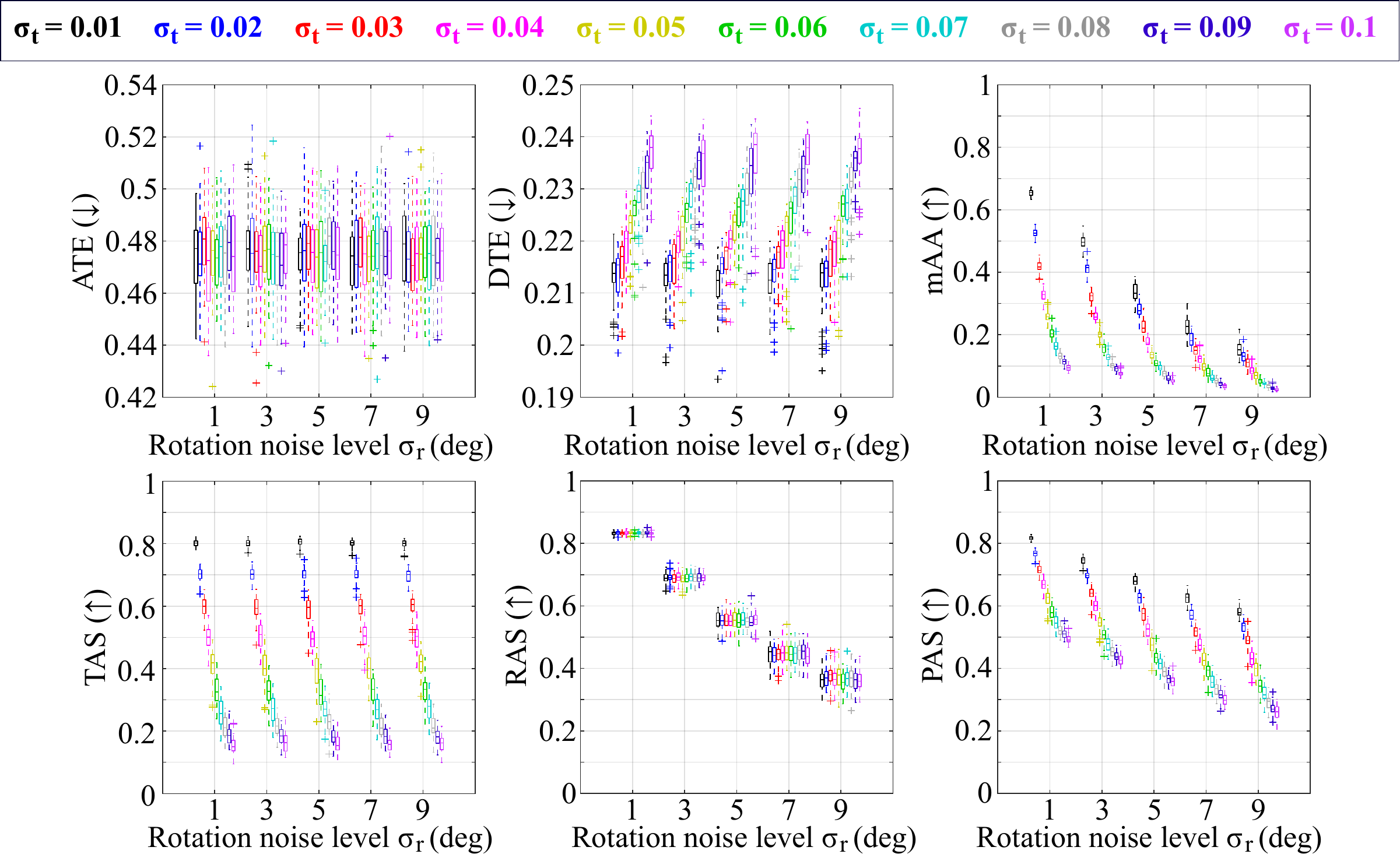}
\caption{\textbf{[Random translations and rotations]} Comparison of the six metrics under different noise levels in translations and rotations.
Due to the presence of outliers, the ATE fails to discern the varying level of noise in both translations and rotations.
On the other hand, the DTE and TAS can discern the varying level of noise in translations, but not in rotations, while the opposite is true for the RAS.
The only two metrics that are responsive to both translation and rotation noise are the mAA and PAS.
Comparing these two in terms of the sensitivity to the noise level, we see that the PAS has more consistent sensitivity across different noise levels.
For instance, the range of the mAA at $\sigma_r=9^\circ$ is 66\% smaller than that at $\sigma_r=1^\circ$, while it is 13\% greater for the PAS.
Also, the range of the mAA at $\sigma_t=0.1$ is 84\% smaller than that at $\sigma_t=0.01$, while it is 14\% greater for the PAS.
}
\label{fig:results_pas}
\end{figure*}

\begin{figure*}[t]
 \centering
 \includegraphics[width=\textwidth]{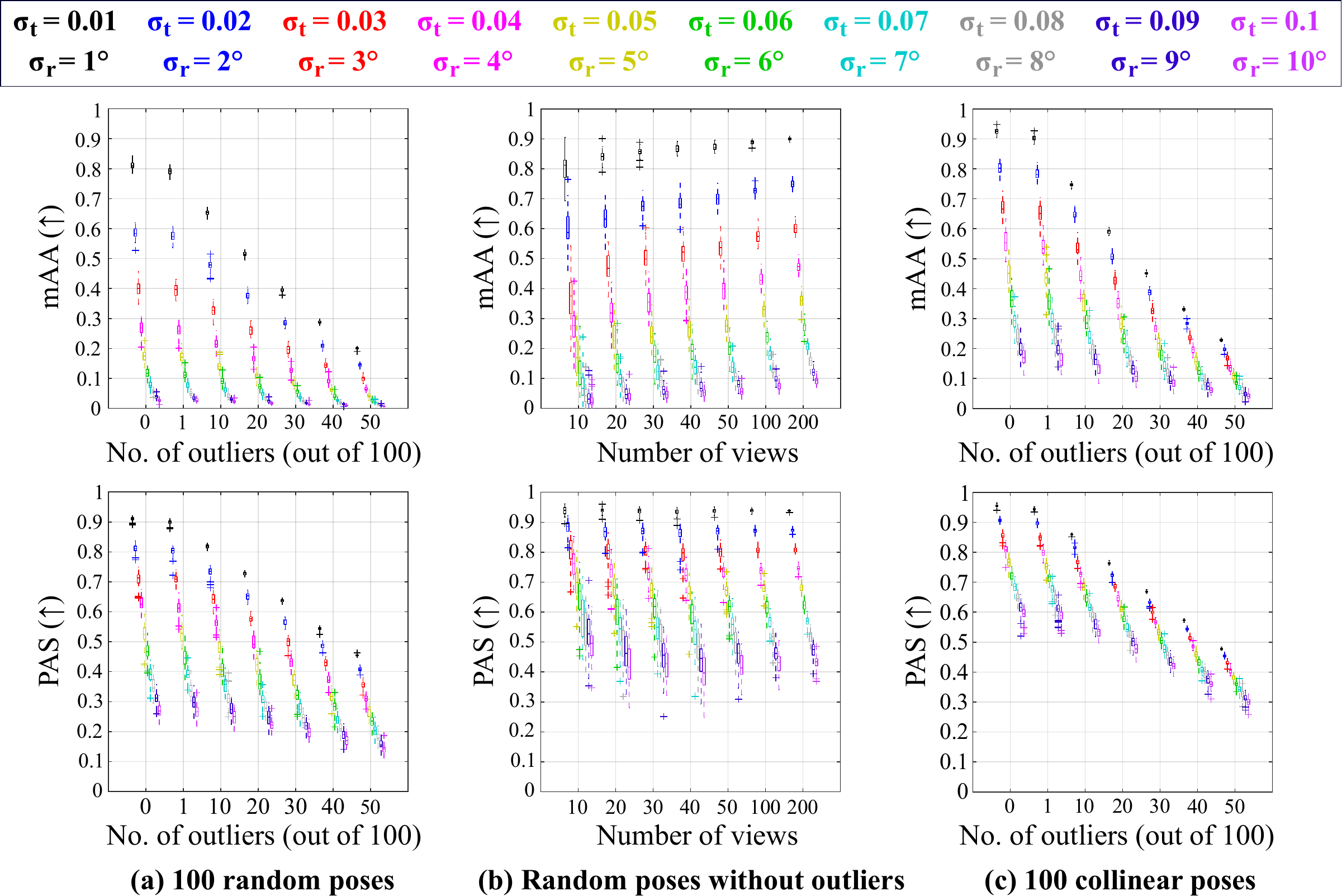}
\caption{\textbf{(a) 1st column:} Comparison of the mAA and PAS for 100 random poses under different noise levels and numbers of outliers.
It shows that the PAS has more consistent sensitivity to both noise level and number of outliers.
For instance, the range of the mAA at 50 outliers is 75\% smaller than that at zero outliers, while it is 50\% smaller for the PAS.
Also, the range of the mAA at $(\sigma_t, \sigma_r)=(0.1, 10^
\circ)$ is 94\% smaller than that at $(\sigma_t, \sigma_r)=(0.01, 1^\circ)$, while it is 55\% smaller for the PAS.
\textbf{(b) 2nd column:} Comparison of the mAA and PAS under different noise levels and numbers of views. 
The mAA tends to favor larger datasets when the noise level is the same.
At moderate noise levels, the PAS is much less sensitive to the change in the number of views than the mAA.
\textbf{(c) 3rd column:} Comparison of the mAA and PAS for 100 collinear poses under different noise levels and numbers of outliers.
Unlike the previous observation in Fig. \ref{fig:results_collinear} where the mAA could barely discern the varying noise level, the mAA in this experiment clearly shows strong discerning power.
This is because this time we varied the noise level in not only translations but also the rotations, and the detection of rotation errors are unaffected by the collinearity.
However, it still shows that the PAS has more consistent sensitivity to both noise level and number of outliers.
For instance, the range of the mAA at 50 outliers is 74\% smaller than that at zero outliers, while it is 59\% smaller for the PAS.
Also, the range of the mAA at $(\sigma_t, \sigma_r)=(0.1, 10^
\circ)$ is 73\% smaller than that at $(\sigma_t, \sigma_r)=(0.01, 1^\circ)$, while it is 24\% smaller for the PAS.
}
\label{fig:results_maa_vs_pas}
\end{figure*}

\subsubsection{Robustness to collinear motion}
\label{subsec:result_collinear}

To evaluate the robustness to collinear camera motion, we modify the simulations in Section \ref{subsec:result_collinear} such that the ground-truth cameras are equally spaced by 1 unit interval along a straight line. 
The results and their interpretation are given in Figure \ref{fig:results_collinear} and its caption, respectively.

\subsubsection{Impact of the trajectory length}
\label{subsec:result_size}

Here, we evaluate the impact of the trajectory length on the translation metrics.
We do this by varying the number of cameras, while keeping the ``density" constant at 10 cameras per unit cubed. 
Specifically, we set $n\in\{10, 20, 30, 40, 50, 100, 200\}$ and randomly place these $n$ cameras inside a cube whose volume is equal to $10n$.
The results of 50 Monte Carlo runs are presented in Figure \ref{fig:results_volume} and commented in its caption.

\subsection{Evaluation of RAS}
\label{sec:result_ras}

In this section, we compare RAS with seven other metrics for camera rotation evaluation:

\begin{itemize}
    \item Median-1 \cite{chatterjee_2018_tpami, lee_2021_cvpr}, Mean-1 \cite{lee_2021_cvpr, lee_2022_cvpr}, RMS-1 \cite{dte}: 
    the median, mean, and root-mean-square (RMS) of the angular errors after aligning the rotations by minimizing the geodesic distances under the $L_1$ norm \cite{hartley_2011_cvpr}.
    \item Median-2 \cite{chatterjee_2018_tpami, lee_2021_cvpr}, Mean-2 \cite{lee_2021_cvpr}, RMS-2 \cite{lee_2022_cvpr}: the counterparts of the previous three, obtained by aligning the rotations using the $L_2$ norm minimization \cite{hartley_2013_ijcv}.
    \item DRE \cite{dte}.
\end{itemize}

To show the impact of the varying noise level and numbers of outliers, we generate 100 random ground-truth rotations.
Then, we obtain the estimated rotations by perturbing them with random rotations with angle $\theta\sim\mathcal{N}(0, \sigma_r^2)$, $\sigma_r\in\{1^\circ, 2^\circ, \cdots, 10^\circ\}$. 
We turn some of the rotations into outliers by replacing them with random rotations.
Figure \ref{fig:results_rotation} presents the results of 50 Monte Carlo runs, in which it can be observed a higher sensitivity of the RAS to noise and outliers than competing baselines.

\subsection{Evaluation of PAS}
\label{sec:result_pas}

\subsubsection{Sensitivity to translation and rotation noise}
In this section, we evaluate the sensitivity to the noise level in both translations and rotations.
We compare the PAS with the ATE \cite{sturm_2012_benchmark}, DTE \cite{dte}, mAA \cite{jin_2021_ijcv}, as well as TAS and RAS.
The experiments are set up as follows:
We generate 100 ground-truth cameras in a unit cube.
Then, we perturb their positions with $\mathcal{N}(0, \sigma_t^2)$, $\sigma_t\in\{0.01, 0.02, \cdots, 0.1\}$ and rotations with $\mathcal{N}(0, \sigma_r^2)$, $\sigma_r\in\{1^\circ, 3^\circ, 5^\circ, 7^\circ, 9^\circ\}$.
We turn 10 out of 100 cameras into outliers using the same method described in Section \ref{subsec:result_outlier}.
The results of 50 Monte Carlo runs are shown in Figure \ref{fig:results_pas}, and the details of the analysis are given in the caption.

\subsubsection{Comparing mAA and PAS}
In this section, we exclusively compare the mAA and PAS across the varying noise noise level in both translation and rotations.
We consider three specific cases:

\begin{enumerate}[label=(\alph*)]
    \item 100 views with random ground-truth poses (\textit{i.e.}, random rotations and random translations) are generated inside a unit cube, some of which are turned into outliers, as described in Section \ref{subsec:result_outlier}.
    \item The number of views (with random ground-truth poses) are varied from 10 to 200, as described in Section \ref{subsec:result_size}. 
    No outliers are present in this experiment.
    \item 100 views with collinear ground-truth poses (\textit{i.e.}, random rotations and collinear translations) are generated, some of which are turned into outliers, as described in Section \ref{subsec:result_collinear}.
\end{enumerate}
Figure \ref{fig:results_maa_vs_pas} shows the results of 50 Monte Carlo runs. Again, the analysis and details are given in the caption of the figure and omitted here to avoid repetition.

\section{Summary of Findings}
\label{sec:summary}

The findings from Section \ref{sec:evaluation} are summarized below:

\begin{enumerate}
    \item When the rotation noise level is fixed and the translation noise level is varied:
    The TAS has stronger discerning power than the ATE and DTE in the presence of outliers. 
    It is also much more robust than the DTE and mAA against the change in the number of views.
    In case of collinear motion, the TAS is the only metric that maintains strong discerning power across the varying amount of noise and outliers.
    \item Among the rotation-only metrics, the RAS has the strongest discerning power across the varying amount of noise and outliers.
    \item When the noise level is varied in either rotations only or translations only, the mAA and PAS can discern it in both cases, while the other metrics fail to do so.  
    Comparing these two metrics, the PAS is much less sensitive than the mAA to the change in the number of views at moderate noise levels.
    Furthermore, the sensitivity of the PAS is more consistent across the varying amount of noise and outliers.
\end{enumerate}

\section{Discussions}
\label{sec:discussion}
\textbf{Which metrics should be used, TAS, RAS or PAS?:}
If the ground-truth data contains only the camera positions, then the TAS is the only option among the three proposed metrics.
If it contains only the rotations, then the RAS is the only option.
If both positions and rotations are known, we recommend reporting either PAS or both TAS and RAS side by side.
The PAS has the advantage that the evaluation can be based on a single metric, which makes it easier to rank the methods and to present the results compactly.
On the other hand, reporting both TAS and RAS side by side has the advantage that the rotation accuracy and the translation accuracy are evaluated separately, which can provide additional insight into the properties of the estimation method. 

\textbf{Limitations:}
Like any other metrics, our proposals also have their own weaknesses.
One of the main weaknesses is that, like the mAA, they respond to outliers much more ``gently" than the ATE and DTE do.
This can be seen as an advantage in terms of robustness against a large number of outliers, but also as a disadvantage if outliers should be highly penalized in the metrics. For example, tracking failures are critical in some real-time applications interacting with the world or a user, such as AR/VR or autonomous robot navigation.
For those applications that need to penalize tracking failures more harshly than the overall increase in the estimation noise, we recommend reporting the ATE (which is extremely harsh against any failure) or DTE (which is moderately harsh).

Another weakness is that the computation of our metrics involves a few heuristics.
One is the choice of the error thresholds in the cumulative frequency histogram (see Fig. \ref{fig:histogram}).
If we choose to use slightly different maximum thresholds, the sensitivity of our metrics will also slightly change.
Another heuristic is the way we combine the TAS and RAS to obtain the PAS in \eqref{eq:pas}.
Equation \eqref{eq:pas} can be seen as an instance of a weighted average between the TAS and RAS, \textit{i.e.}, $\alpha\mathrm{TAS}+(1-\alpha)\mathrm{RAS}$, with $0\leq\alpha\leq1$.
The greater the weight $\alpha$, the greater the contribution of the TAS relative to the RAS.
In our definition of the PAS \eqref{eq:pas}, we simply chose to set $\alpha=0.5$.

\section{Conclusions}
\label{sec:conclusion}
We proposed three novel metrics for multiview pose accuracy evaluation.
The TAS and RAS evaluate the accuracy of the camera translations and rotations, respectively.
The PAS is their average, representing the accuracy of the 6 degrees-of-freedom poses as a whole.
The main idea behind our proposal is to first robustly align the estimated camera translations/rotations to the ground truth, and then generate the cumulative frequency histograms of the errors, from which we obtain the truncated area below certain thresholds.
Through extensive simulations, we demonstrated the practical advantages of our metrics over the existing ones: namely, greater robustness to outliers and collinearity, reduced sensitivity to the change in trajectory length, and better consistency across the varying amount of noise and outliers.

\end{document}